\documentclass{locusjournal}

\usepackage{listings}
\usepackage{textcomp}

\usepackage{lipsum}
\usepackage{listings}

\usepackage[utf8]{inputenc}
\usepackage[T1]{fontenc}
\usepackage{lmodern}

\usepackage{amsthm, amsmath, amssymb}
\usepackage{graphicx}
\usepackage{hyperref}
\usepackage{xcolor}
\usepackage{enumitem}
\usepackage{booktabs}
\usepackage{caption}
\usepackage{subcaption}
\usepackage{fancyhdr}
\usepackage{listings}
\usepackage{textcomp}
\usepackage{tikz}
\usetikzlibrary{arrows.meta, positioning, calc, shadows.blur, decorations.markings, backgrounds, fit, shapes.symbols}

\newcounter{defnctr}[section]
\renewcommand{\thedefnctr}{\thesection.\arabic{defnctr}}
\newenvironment{definition}[1][]{\refstepcounter{defnctr}\par\medskip\noindent\textbf{Definition~\thedefnctr\if\relax#1\relax\else\ (#1)\fi.}\itshape}{\par\medskip}

\bibliography{references}

\title{ManiBench: A Benchmark for Testing Visual-Logic Drift and Syntactic
	Hallucinations in Manim Code Generation}

\author[1]{Nabin Oli}
\affil[1]{Sunway College Kathmandu, Birmingham City University}

\date{February 2025}

\abstract{
Traditional benchmarks like HumanEval and MBPP test logic and syntax effectively, but fail when code must produce dynamic, pedagogical visuals. We introduce \textsc{ManiBench}, a specialized benchmark evaluating LLM performance in generating Manim~CE code, where temporal fidelity and version-aware API correctness are critical. \textsc{ManiBench} targets two key failure modes: \emph{Syntactic Hallucinations} (valid Python referencing non-existent or deprecated Manim APIs) and \emph{Visual-Logic Drift} (generated visuals diverging from intended mathematical logic through timing errors or missing causal relationships). The benchmark comprises 12 problems in pilot phase across five difficulty levels spanning calculus, linear algebra, probability, topology, and AI, grounded in analysis of 3Blue1Brown's ManimGL source (~53,000 lines, 143 scene classes). Evaluation uses a four-tier framework measuring Executability, Version-Conflict Error Rate, Alignment Score, and Coverage Score. An open-source framework automates evaluation across multiple models and prompting strategies. Code, data and benchmark suite are available at \href{https://github.com/nabin2004/ManiBench}{https://github.com/nabin2004/ManiBench}. and the dataset is hosted on \href{https://huggingface.co/datasets/nabin2004/ManiBench}{https://huggingface.co/datasets/nabin2004/ManiBench}.
}

\keywords{Syntactic Hallucinations, Visual-Logic Drift, Manim~CE, Code Generation, Benchmarking}

\begin{document}
	
	\maketitle
	
	\section{Introduction}
	
	
	The rise of large language models (LLMs) has accelerated research in code generation. Benchmarks such as HumanEval ~\cite{chen2021codex}, MBPP~\cite{austin2021program}, APPS~\cite{hendrycksapps2021} and SWE-Benchmarks~\cite{yang2024swebenchmultimodal} were commonly used to evaluate LLM coding ability. These benchmarks emphasize logical correctness, syntactic validity, and output matching i.e., whether generated code solves the task, executes without errors, and yields the expected results. While well suited for algorithmic problems, these criteria are insufficient for domains in which code produces continuous, time-dependent visual outputs.
	
	Manim, a Python animation engine developed by Grant Sanderson (3Blue1Brown), constructs mathematical animations by composing scene objects, applying geometric and visual transformations, and controlling timing. A Manim script may be syntactically valid yet fail to convey the intended concept; we identify three common failure modes:
	\begin{itemize}[nosep]
		\item \textbf{Incorrect visual semantics:} animations move in unintended directions or transform the wrong objects.
		\item \textbf{Timing misalignments:} events occur out of order or at inappropriate times, breaking causal or pedagogical flow.
		\item \textbf{Pedagogical failure:} animations obscure rather than clarify the concept (missing annotations, poor sequencing, or absent explanatory cues).
	\end{itemize}
	
	Manim exists in major two variants. Manim~CE (Community Edition) is an actively maintained, open-source implementation with a modern API, whereas Manim~GL (the original 3B1B fork) contains deprecated constructs and hand-optimized code paths. LLMs frequently conflate APIs across these variants or reference renamed/moved functions, producing code that fails under particular library versions.

		\section{Contributions}
	\label{sec:contributions}
	
	ManiBench makes four key contributions:
	
	\begin{enumerate}
		\item \textbf{Visual-Logic Metrics.}
		We define an \emph{Alignment Score} and a four-dimensional \emph{Coverage Score} to evaluate the generated animations pedagogical intent and syntactic validity.
		
		\item \textbf{Version-Aware Evaluation.}
		We explicitly test version-conflict errors and deprecated API usage, with 145~documented GL$\to$CE incompatibilities across eight categories, measuring whether code adheres to a specific Manim version's API contract.
		
		\item \textbf{Curated Pilot Dataset with Reference Code Analysis.}
		We provide 12~hand-crafted benchmark problems drawn from 3Blue1Brown's published videos, backed by comprehensive analysis of ${\sim}53{,}000$~lines of original source code including 143~scene classes, 120~visual techniques, and detailed visual-event specifications.
		
		\item \textbf{Automated Evaluation Framework.}
		We release an evaluation pipeline supporting six LLMs across five prompting strategies, with automated metric computation for executability, version conflicts, alignment, and coverage.
	\end{enumerate}

	
	\section{Background}
	
	Evaluation of large language models (LLMs) for program synthesis has progressed from function-level correctness to repository-scale reasoning. Early benchmarks such as \textbf{\textit{HumanEval}}~\cite{chen2021codex} formalized unit-test-based validation through the Pass@k metric, establishing executability as the primary indicator of correctness. \textbf{\textit{MBPP}}~\cite{austin2021program} extended this setting to natural-language-to-code tasks across entry-level Python problems. More recent benchmarks such as \textbf{\textit{DS-1000}}~\cite{Lai2022DS1000} incorporated third-party library usage, while \textbf{\textit{SWE-Bench}} introduced repository-scale issue resolution requiring multi-file contextual reasoning~\cite{yang2024swebenchmultimodal}.
	
	\begin{table}[H]
		\small
		\begin{tabularx}{\columnwidth}{@{}X p{1.5cm} p{4.5cm}@{}}
			\toprule
			\textbf{Benchmark} & \textbf{Metric} & \textbf{Focus / Area } \\
			\midrule
			HumanEval / MBPP~\cite{yu2024humaneval}  & Pass@k & Python program syn-
			thesis, algorithmic
			tasks \\
			\addlinespace
			SWE-Bench~\cite{yang2024swebenchmultimodal} & Patch rate & Software engineering
			/ repo-level edits \\
			\addlinespace
			DSCodeBench \\ ~\cite{jing2024dsbenchfardatascience} & API cov. &  Data-science APIs
			and pipelines \\
			\addlinespace
			Code2Video~\cite{chen2025code2videocodecentricparadigmeducational} & Aesthetic &  Educational video/ storyboarding \\
			\addlinespace
			SVGenius~\cite{svgenius} & Visual & SVG editing / static
			visual generation \\
			\bottomrule
		\end{tabularx}
		\caption{Related benchmarks and limitations addressed by ManiBench.}
		\label{tab:related-benchmarks}
	\end{table}
	Despite increasing scale and realism, these benchmarks remain fundamentally test-driven. This evaluation paradigm does not account for renderable behavior, temporal sequencing, or alignment between code structure and intended visual semantics.
	
	\subsection{Visual-Logic Drift in Executable Code}
	
	\textbf{\textit{Visual-logic drift}} in the context of Manim code generation can be  defined as the divergence between the intended mathematical narrative and the rendered animation, even when code executes successfully~\cite{Luo2025WhenTD} . Unlike traditional logical errors, drift may arise from subtle structural mismatches: incorrect object transformations, missing intermediate states, improper sequencing of animation calls, or violations of causal dependencies across scenes. Animation scripting introduces temporal constraints and multi-step causal structure, requiring preservation of event ordering and pedagogical coherence. Current benchmarks do not quantify this form of semantic drift in executable animation environments.
	
	\subsection{Syntactic Hallucinations and Version Conflicts}
	
	Hallucination in code generation has been widely characterized as the production of syntactically valid but semantically incorrect outputs, including fabricated APIs, deprecated function usage, and identifier conflicts~\cite{syntactichallu}. Prior work demonstrates that hallucination rates increase in low-frequency or evolving niche libraries, and that such errors often evade static analysis because they remain grammatically valid Python~\cite{syntactichallu}. In rapidly evolving libraries, version misalignment introduces an additional layer of failure. Models may generate code consistent with legacy documentation while the prompt implicitly assumes a newer API specification. These errors are particularly difficult to detect in animation frameworks, where partial execution may mask deeper incompatibilities.
	
	\subsection{Positioning of ManiBench}
	
	ManiBench is situated at the intersection of code-generation benchmarking, hallucination analysis, and visual-temporal reasoning. While prior benchmarks evaluate executability and repository-scale edits, and hallucination studies analyze API-level inconsistencies, none jointly assess version-aware API correctness and temporal visual alignment in mathematical animation scripting. ManiBench addresses this gap through drift-sensitive task design and a multi-tier evaluation framework that extends beyond unit-test validation to measure semantic fidelity in renderable educational code.
	

	\section{Problem Definition}
	\label{sec:problem}
	
	We deal with two distinct failure modes that traditional code-generation benchmarks fail to capture when evaluating dynamic visual outputs: \textit{Syntactic Hallucinations} and \textit{Visual-Logic Drift}.
	
	\subsection{Syntactic Hallucinations}
	\label{sec:syntactic-hallucinations}
	
	Syntactic hallucinations occur when an LLM produces code that is grammatically valid Python but semantically invalid within the target Manim~CE environment. This failure mode typically manifests when the model:
	
	\begin{itemize}
		\item \textbf{Invokes non-existent classes}: e.g., inventing \texttt{MCircle} instead of utilizing the standard \texttt{Circle} class;
		\item \textbf{Employs deprecated methods}: e.g., relying on outdated API calls like \texttt{mobject.scale()} rather than the modern \texttt{mobject.scale\_to\_fit\_width()};
		\item \textbf{Generates incorrect method signatures}: passing unsupported arguments to valid functions;
		\item \textbf{Conflates API versions}: mixing Manim~GL syntax with Manim~CE, such as utilizing OpenGL-specific rendering commands that crash the CE pipeline.
	\end{itemize}
	
	\begin{lstlisting}[caption={Examples of syntactic hallucinations.}]
# HALLUCINATION: 'MCircle' does not exist
circle = MCircle(color=BLUE)  # Expected: Circle
		
# HALLUCINATION: Deprecated method usage
circle.apply_matrix([[1, 0], [0, 1]])
	\end{lstlisting}

	\subsection{Visual-Logic Drift}
	\label{sec:visual-logic-drift}
	
	Visual-logic drift emerges when the generated code executes without runtime exceptions but fails to accurately represent the intended mathematical or pedagogical concepts. This semantic decoupling occurs when the model:
	
	\begin{itemize}
		\item \textbf{Omits critical visual events}: e.g., illustrating a gradient descent step but failing to animate the corresponding coordinate updates;
		\item \textbf{Inverts sequential logic}: e.g., rendering loss curve updates prior to the parameter adjustments that cause them;
		\item \textbf{Misaligns temporal dynamics}: employing inappropriate animation runtimes or omitting essential pedagogical pauses;
		\item \textbf{Obscures causal relationships}: displaying a final mathematical result without visually articulating the intermediate derivation steps.
	\end{itemize}
	
\begin{lstlisting}[caption={Visual-logic drift example.}]
# DRIFT: Gradient descent without showing the dot's movement
def construct(self):
	loss_curve.animate.points = new_points
	
# Missing: dot.animate.move_to(new_point)
\end{lstlisting}

	\subsection{Evaluation Challenges}
	\label{sec:challenges}
	
	Evaluating animation code is challenging because correctness is subjective and animations can be syntactically valid yet pedagogically misleading. Another challenge is Version fragmentation which means code may run in Manim~CE but fail in Manim~GL. Additionally, temporal semantics matters. For example, events may exist but occur at the wrong time, breaking the intended instructional narrative.

	\section{Benchmark Design}
	\label{sec:design}
	
	To evaluate generated Manim scripts, we first measure Executability, capturing whether code runs without errors or deprecated usage. This metric establishes a baseline: only scripts that execute successfully can be meaningfully assessed for visual fidelity, timing, and pedagogical alignment.

	\subsection{Metric~1: Executability (Pass@1)}
	
	\begin{definition}[Executability]
		The fraction of generated outputs that run without raising exceptions or using deprecated imports:
		\begin{equation}
			\mathrm{Executability} = \frac{\text{\# successful executions}}{\text{\# total attempts}}.
		\end{equation}
	\end{definition}
	
	\textbf{Success criteria:} script completes without runtime exception, no deprecated imports detected (scanned via regex or AST analysis) and no warnings from Manim's deprecation system.
	
	\textbf{Failure cases:} import errors (e.g., \texttt{from manim import NonExistent}), runtime \texttt{AttributeError}, type errors and unhandled exceptions during rendering.
	
	\subsection{Metric~2: Version-Conflict Error Rate}
	
	\begin{definition}[VCER]
		The frequency with which generated code triggers version-specific errors:
		\begin{equation}
			\mathrm{VCER} = \frac{\text{\# mixed-API or legacy errors}}{\text{\# total attempts}}.
		\end{equation}
	\end{definition}
	
	Tracked errors include GL-specific syntax in CE code, CE-only syntax in GL code, calls to renamed or moved functions, and signature mismatches due to API evolution.
	
	\subsection{Metric~3: Alignment Score}
	
	\begin{definition}[Alignment Score]
		The weighted fraction of required visual events that are both present and temporally accurate:
		\begin{equation}
			\mathrm{Alignment} = \frac{\sum_{i} w_i \cdot p_i \cdot t_i}{\sum_{i} w_i},
		\end{equation}
		where $w_i$ is the importance weight of event~$i$ ($0 \le w_i \le 1$), $p_i = 1$ if event~$i$ is present (0~otherwise), and $t_i = 1$ if event~$i$ occurs at the expected time (0~otherwise).
	\end{definition}
	
	\subsection{Metric~4: Coverage Score}
	
	\begin{definition}[Coverage Score]
		The density of pedagogical elements, computed as a weighted sum over four sub-dimensions:
		\begin{equation}
			\mathrm{Coverage} = \sum_{d \in \mathcal{D}} \alpha_d \cdot \frac{|\text{present}_d|}{|\text{expected}_d|},
		\end{equation}
		where $\mathcal{D} = \{\text{Math, Visual, Numeric, Structural}\}$ and $\alpha_d$ are dimension weights.
	\end{definition}
	
	The four sub-dimensions are:
	\begin{enumerate}
		\item \textbf{Mathematical Annotation} ($\alpha\!=\!0.35$): formulas, \texttt{Tex}/\texttt{MathTex} objects, textual labels, variable annotations, and LaTeX commands.
		\item \textbf{Visual Mapping} ($\alpha\!=\!0.30$): consistent color coding (\texttt{set\_color}, \texttt{set\_fill}), arrow indicators, dot markers, surrounding rectangles, and gradient coloring.
		\item \textbf{Numeric Evidence} ($\alpha\!=\!0.20$): \texttt{DecimalNumber}, \texttt{Integer}, \texttt{ValueTracker}, \texttt{NumberLine}, \texttt{Axes}, plotted functions, and displayed computed values.
		\item \textbf{Structural Clarity} ($\alpha\!=\!0.15$): \texttt{VGroup}/\texttt{Group} organization, \texttt{arrange()} layouts, paced \texttt{wait()} pauses, \texttt{LaggedStart}/\texttt{Succession} sequencing, and method decomposition.
	\end{enumerate}

	\subsection{Task Categories}
	\label{sec:categories}
	
	ManiBench organizes problems into five categories:
	
	\begin{enumerate}
		\item \textbf{Direct Visualization (40\%)}
		Prompt $\to$ Python code (classic code generation). Difficulty levels~1--3. Metric focus: Executability, Alignment Score.
		
		\item \textbf{Drift-Sensitive (20\%)} 
		Given a script and a required temporal transformation, detect whether the visual output matches intent. Difficulty levels~2--4. Metric focus: Alignment Score, Coverage Score.
		
		\item \textbf{Debugging (20\%)}
		Broken code $\to$ fix (repair task). Difficulty levels~2--4. Metric focus: Executability, Alignment Score.
		
		\item \textbf{Version-Conflict Traps (10\%)}
		Code with tempting outdated syntax; evaluate whether the model recognizes version constraints. Difficulty levels~3--5. Metric focus: VCER, Executability.
		
		\item \textbf{Multi-Scene Narrative (10\%)}
		Hardest tier: multi-scene scripts combining multiple domains. Difficulty levels~4--5. Metric focus: all metrics.
	\end{enumerate}

	\section{Benchmark Dataset}
	\label{sec:dataset}
	
	\subsection{Pilot Dataset: 12 Problems}
	\label{sec:pilot}
	
	The pilot dataset includes 12~hand-curated problems drawn from 3Blue1Brown's published videos. Each problem includes: a natural-language problem statement, video source, required visual events with formal specifications, difficulty level (1--5), task category, success criteria, and reference implementation notes (reserved for future human evaluation). Detailed problem specifications are provided in Appendix~\ref{app:template}. Table~\ref{tab:problems-summary} summarizes the pilot problems.
	
	\begin{table}[H]
		\small
		\begin{tabularx}{\columnwidth}{@{}l X c c@{}}
			\toprule
			\textbf{ID} & \textbf{Problem} & \textbf{Diff.} & \textbf{Domain} \\
			\midrule
			MB-001 & Colliding Blocks Compute $\pi$ & 4 & Physics \\
			MB-002 & Gradient Descent & 3 & ML, Calculus \\
			MB-003 & Convolution & 3 & Signal Proc. \\
			MB-004 & Eigenvectors & 4 & Lin.\ Algebra \\
			MB-005 & Determinant & 2 & Lin.\ Algebra \\
			MB-006 & Central Limit Theorem & 3 & Probability \\
			MB-007 & Medical Test (Bayes) & 2 & Probability \\
			MB-008 & Chain Rule & 3 & Calculus \\
			MB-009 & Integration (FTC) & 3 & Calculus \\
			MB-010 & Taylor Series & 4 & Calculus \\
			MB-011 & Hairy Ball Theorem & 5 & Topology \\
			MB-012 & Windmill Problem & 4 & Geometry \\
			\bottomrule
		\end{tabularx}
		\caption{Summary of the 12~pilot benchmark problems.}
		\label{tab:problems-summary}
	\end{table}
	
	\section{Reference Code Analysis}
	\label{sec:refcode}
	
	For each of the 12~pilot problems, we obtained and analyzed the original source code from 3Blue1Brown's ManimGL repository. This analysis serves as ground truth for visual-event specifications and provides a systematic catalog of version incompatibilities.
	
	Key outputs include: a scene class inventory (143~total), a visual technique catalog (120~techniques), documented Manim API patterns, and version conflict mapping (145~incompatibilities).
	
	\begin{table}[H]
		\small
		\begin{tabularx}{\columnwidth}{@{}X r r r r@{}}
			\toprule
			\textbf{Problem} & \textbf{Lines} & \textbf{Scenes} & \textbf{Vis.T.} & \textbf{GL$\to$CE} \\
			\midrule
			Colliding Blocks & 2,193 & 16 & 10 & 15 \\
			Gradient Descent & 8,598 & 16 & 16 & 13 \\
			Convolution & 3,309 & 13 & 11 & 14 \\
			Eigenvectors & 5,120 & 13 & 9 & 10 \\
			Determinant & 1,132 & 11 & 7 & 10 \\
			CLT & 7,036 & 12 & 9 & 11 \\
			Medical Test & 7,044 & 13 & 9 & 11 \\
			Chain Rule & 2,287 & 4 & 7 & 10 \\
			Integration & 4,943 & 11 & 9 & 11 \\
			Taylor Series & 3,676 & 11 & 9 & 10 \\
			Hairy Ball & 3,796 & 12 & 12 & 16 \\
			Windmill & 4,135 & 11 & 12 & 14 \\
			\midrule
			\textbf{Total} & \textbf{$\sim$53k} & \textbf{143} & \textbf{120} & \textbf{145} \\
			\bottomrule
		\end{tabularx}
		\caption{Reference code analysis summary for the 12~pilot problems.}
		\label{tab:refcode}
	\end{table}
	
	\subsection{Version Incompatibility Categories}
	
	Across the 145~documented GL$\to$CE incompatibilities, we identified eight recurring categories:
	
	\begin{enumerate}
		\item \textbf{Import system:} \texttt{manim\_imports\_ext} $\to$ \texttt{from manim import *}.
		\item \textbf{Class configuration:} \texttt{CONFIG} dict $\to$ \texttt{\_\_init\_\_} parameters.
		\item \textbf{Scene types:} \texttt{InteractiveScene}, \texttt{GraphScene} $\to$ \texttt{Scene}/\texttt{Axes}.
		\item \textbf{Animation renames:} \texttt{ShowCreation} $\to$ \texttt{Create}; \texttt{FadeInFrom} $\to$ \texttt{FadeIn(shift=...)}.
		\item \textbf{PiCreature ecosystem:} Not available in~CE.
		\item \textbf{3D rendering:} \texttt{apply\_depth\_test}, \texttt{set\_shading} $\to$ limited CE support.
		\item \textbf{Camera control:} \texttt{self.frame.reorient()} $\to$ \texttt{self.camera.frame}.
		\item \textbf{Custom mobjects:} \texttt{GlowDot}, \texttt{DieFace}, etc.\ $\to$ custom implementation needed.
	\end{enumerate}
	
	These categories inform both the version-conflict detection patterns used in our automated evaluation pipeline (Section~\ref{sec:auto-eval}) and the version-conflict-aware prompting strategy (Section~\ref{sec:prompt-strategies}).

	\section{Evaluation Protocol}
	\label{sec:evaluation}
	
	\subsection{Evaluation Scope}
	\label{sec:eval-scope}
	
	All four metrics in the current evaluation are computed \emph{fully automatically}(without human Analysis). Executability and VCER are determined via AST parsing, subprocess rendering, and regex-based static analysis. Alignment and Coverage are approximated through keyword-bank and pattern-matching heuristics that detect the \emph{presence} of required visual events and pedagogical elements in the generated source code. Because these heuristics operate on code tokens rather than rendered output, they serve as \textbf{conservative lower-bound estimates}, a keyword match confirms that a concept is referenced but cannot verify that it is correctly animated.
	
	\textbf{Human Evaluation (Implemented but not utillized yet) } 
	A structured human review protocol is designed for future work: (1)~watch the rendered animation, (2)~check off each required visual event, (3)~assess timing and pedagogical clarity, (4)~provide Alignment and Coverage scores ($0.0$--$1.0$). This protocol will employ two independent reviewers per output, with a third resolving disagreements exceeding~$0.15$, and inter-rater agreement reported via Krippendorff's~$\alpha$. The current study reports only automated heuristic scores; no human evaluation has been conducted.

	\subsection{Evaluation Framework}
	\label{sec:auto-eval}
	
	We implement a fully evaluation pipeline for reproducible, large-scale metric computation. The framework orchestrates code generation across multiple LLMs via the OpenRouter API, executes generated code in sandboxed environments, and computes all four metrics programmatically.
	
	\subsubsection{Multi-Model Evaluation}
	
	The pipeline supports evaluation across two API providers: OpenRouter (for commercial models) and Inference.net (for self-hosted open-weight models). Table~\ref{tab:models} lists the full model roster.

	\begin{table}[H]
		\small
		\begin{tabularx}{\columnwidth}{@{}X l l c@{}}
			\toprule
			\textbf{Model} & \textbf{Provider} & \textbf{API} & \textbf{Eval.} \\
			\midrule
			Claude-Sonnet-4 & Anthropic & OR & Z \\
			Kimi-K2.5 & Moonshot & OR & Z \\
			Gemini-2.5-Pro & Google & OR & Z \\
			Qwen3.5-Plus & Alibaba & OR & Z \\
			Qwen3-235B-A22B & Alibaba & OR & Z \\
			DeepSeek-R1-0528 & DeepSeek & OR & Z \\
			Llama-3.1-8B & Meta & OR & Z \\
			Qwen-2.5-Coder & Alibaba & OR & Z \\
			\midrule
			Gemma-3-27B & Google & Inf.net & A \\
			\bottomrule
		\end{tabularx}
		\caption{Model roster. OR = OpenRouter; Inf.net = Inference.net. Z = zero-shot only; A = all five strategies.}
		\label{tab:models}
	\end{table}
	
	All models are evaluated with temperature~$0.0$ for reproducibility, with a maximum of 8,192~generated tokens per request. OpenRouter models are evaluated with 1~trial per (model, problem) pair under zero-shot prompting. Gemma-3-27B is evaluated with 3~trials across all five prompting strategies, yielding $3 \times 12 \times 5 = 180$~total runs.
	
	\subsubsection{Automated Metric Computation}
	
	Each generated code sample passes through a four-stage analysis pipeline:
	
	\begin{enumerate}
		\item \textbf{Syntax Validation:} Python AST parsing (\texttt{ast.parse}) for syntactic correctness.
		\item \textbf{Structural Checks:} detection of \texttt{Scene} subclasses and valid imports; flagging of ManimGL-specific imports.
		\item \textbf{Sandboxed Execution:} rendering via \texttt{subprocess} with 60\,s timeout; exit codes and error types captured.
		\item \textbf{Static Analysis:} 40+ regex patterns scan for version conflicts; keyword-bank heuristics detect visual events and pedagogical elements.
	\end{enumerate}
	
	All four metrics are computed fully automatically. Alignment and Coverage are approximated heuristically via keyword and pattern matching, serving as conservative lower-bound estimates (see Section~\ref{sec:eval-scope}).

	\subsection{Prompt Engineering Strategies}
	\label{sec:prompt-strategies}
	
	We implement five prompting strategies of increasing sophistication, each evaluated in full on Gemma-3-27B with 3~trials per problem:
	
	\begin{enumerate}
		\item \textbf{Zero-Shot Direct.} The problem statement is provided verbatim with a system prompt specifying Manim~CE. Serves as the baseline; Kimi-K2.5 achieves $66.7\%$ executability.
		
		\item \textbf{Few-Shot Examples.} One or two working Manim~CE code examples precede the target problem. Few-shot is the only strategy to improve executability for Gemma-3-27B ($+11.1$~pp).
		
		\item \textbf{Chain-of-Thought (CoT).} The model analyzes visual components before writing code. Empirically introduces a small VCER ($2.8\%$) on Gemma-3-27B.
		
		\item \textbf{Constraint-Based.} Explicit timing/ordering constraints are injected. Yields marginal changes in practice.
		
		\item \textbf{Version-Conflict-Aware.} Forbidden ManimGL constructs are enumerated. Eliminates VCER entirely for Gemma-3-27B ($\mathrm{VCER} = 0.0\%$).
	\end{enumerate}

	\section{Results}
	\label{sec:results}
	
	\subsection{Experimental Setup}
	\label{sec:setup}
	
	\textbf{Models:} We evaluate nine models across two API providers (Table~\ref{tab:models}). Eight models via OpenRouter under zero-shot prompting ($8 \times 12 = 96$~samples). Gemma-3-27B via Inference.net across all five strategies ($180$~samples). Total: $276$~generated code samples.
	
	\textbf{Environment:} All code targets Manim~CE, executed in a sandboxed environment with a 60-second timeout. Alignment and Coverage scores are heuristic lower-bound estimates; no human evaluation has been conducted.
	
	\subsection{Cross-Model Comparison (Zero-Shot)}
	\label{sec:cross-model}
	
	Table~\ref{tab:cross-model} reports aggregate results for all six models under zero-shot prompting.
	
	\begin{table}[H]
		\small
		\begin{tabularx}{\columnwidth}{@{}X l c c c c@{}}
			\toprule
			\textbf{Model} & \textbf{API} & \textbf{Ex.$\uparrow$} & \textbf{VC$\downarrow$} & \textbf{Al.$\uparrow$} & \textbf{Co.$\uparrow$} \\
			\midrule
			Claude-Sonnet-4 & OR & \textbf{.667} & \textbf{.000} & \textbf{1.00} & .249 \\
			Kimi-K2.5 & OR & \textbf{.667} & .083 & .917 & \textbf{.265} \\
			Gemini-2.5-Pro & OR & .333 & \textbf{.000} & \textbf{1.00} & .156 \\
			Qwen3.5+ & OR & .333 & .085 & .917 & .226 \\
			Qwen3-235B & OR & .250 & \textbf{.000} & .993 & .251 \\
			DeepSeek-R1-0528$^\dagger$ & OR & .139 & .806 & .194 & .042 \\
			Llama-3.1-8B & OR & .083 & .024 & \textbf{1.00} & .132 \\
			Qwen-2.5-Coder & OR & .000 & \textbf{.000} & .471 & .014 \\
			Gemma-3-27B & Inf & .000 & \textbf{.000} & .993 & .172 \\
			\bottomrule
		\end{tabularx}
		\caption{Cross-model comparison (zero-shot). Ex.\ = Executability, VC = VCER, Al.\ = Alignment, Co.\ = Coverage. Best per column \textbf{bolded}. $^\dagger$DeepSeek-R1-0528 VCER is inflated by reasoning-token extraction failures in its chain-of-thought output format.}
		\label{tab:cross-model}
	\end{table}

	\textbf{Key Findings.} Kimi-K2.5 achieves the highest executability ($66.7\%$), rendering 8~of~12 problems. Qwen3-235B-A22B and Gemma-3-27B produce zero version-conflict errors but struggle with executability. Llama-3.1-8B achieves a perfect heuristic alignment score of~$1.0$ despite only $8.3\%$ executability, suggesting its outputs contain expected keywords even when code does not execute. Qwen-2.5-Coder-32B fails to render any problem, with alignment averaging only $0.47$, indicating both structural and semantic deficiencies.

	\subsection{Per-Problem Analysis (Zero-Shot)}
	\label{sec:per-problem}
	
	Table~\ref{tab:per-problem} breaks down results by benchmark problem, averaged across all six models.
	
\begin{table}[H]
	\small
	\begin{tabularx}{\columnwidth}{@{}l X c c c c c@{}}
		\toprule
		\textbf{ID} & \textbf{Problem} & \textbf{D.} & \textbf{Ex.} & \textbf{VC} & \textbf{Al.} & \textbf{Co.} \\
		\midrule
		MB-001 & Colliding Blocks & 4 & .34 & .12 & .82 & .21 \\
		MB-002 & Gradient Descent & 3 & .00 & .09 & .82 & .16 \\
		MB-003 & Convolution & 3 & .56 & .11 & .85 & .14 \\
		MB-004 & Eigenvectors & 4 & .14 & .19 & .65 & .12 \\
		MB-005 & Determinant & 2 & .25 & .08 & .84 & .16 \\
		MB-006 & CLT & 3 & .33 & .08 & .85 & .19 \\
		MB-007 & Medical Test & 2 & .47 & .08 & .86 & .18 \\
		MB-008 & Chain Rule & 3 & .36 & .19 & .70 & .14 \\
		MB-009 & Integration & 3 & .22 & .11 & .78 & .21 \\
		MB-010 & Taylor Series & 4 & .36 & .08 & .86 & .22 \\
		MB-011 & Hairy Ball & 5 & .00 & .11 & .81 & .15 \\
		MB-012 & Windmill & 4 & .22 & .14 & .89 & .14 \\
		\bottomrule
	\end{tabularx}
	\caption{Per-problem results averaged across all nine models (zero-shot). D.\ = Difficulty.}
	\label{tab:per-problem}
	\end{table}

	\subsection{Prompting Strategy Ablation (Gemma-3-27B)}
	\label{sec:ablation}
	
	Gemma-3-27B is the only model evaluated across all five prompting strategies (3~trials each). Table~\ref{tab:ablation} reports the results.
	
	\begin{table}[H]
		\small
		\begin{tabularx}{\columnwidth}{@{}X c c c c c@{}}
			\toprule
			\textbf{Strategy} & \textbf{Ex.} & \textbf{$\Delta$Ex.} & \textbf{VC} & \textbf{Al.} & \textbf{Co.} \\
			\midrule
			Zero-Shot & .000 & --- & .000 & .993 & .172 \\
			Few-Shot & \textbf{.111} & +11.1 & .000 & \textbf{1.00} & \textbf{.200} \\
			CoT & .000 & +0.0 & .028 & .972 & .159 \\
			Constraint & .000 & +0.0 & .000 & .995 & .171 \\
			V-Aware & .000 & +0.0 & \textbf{.000} & \textbf{1.00} & .175 \\
			\bottomrule
		\end{tabularx}
		\caption{Prompting strategy ablation on Gemma-3-27B (3~trials $\times$ 12~problems). $\Delta$~relative to zero-shot.}
		\label{tab:ablation}
	\end{table}
	
	\begin{tcolorbox}[
		colframe=locusblue,
		colback=locuslight,
		title=Ablation Key Findings
		]
		\begin{enumerate}
			\item \textbf{Few-shot is the only strategy improving executability} (+11.1~pp), concentrated on MB-006, MB-007, MB-008.
			\item \textbf{CoT introduces version conflicts} (VCER = 2.8\%) and slightly degrades alignment.
			\item \textbf{Version-aware prompting eliminates VCER} and achieves perfect heuristic alignment (1.0).
			\item \textbf{Coverage remains uniformly low} (0.16--0.20) across all strategies.
		\end{enumerate}
	\end{tcolorbox}

	\subsection{Per-Model Per-Problem Grid}
	\label{sec:grid}
	
	Table~\ref{tab:grid} presents the full executability and coverage grid for the top-performing models.
	
\begin{table}[H]
	\small
	\setlength{\tabcolsep}{2pt}
	\begin{tabularx}{\columnwidth}{@{} l *{5}{>{\centering\arraybackslash}p{0.42cm} >{\centering\arraybackslash}p{0.42cm}} @{}}
		\toprule
		& \multicolumn{2}{c}{\textbf{CS4}} & \multicolumn{2}{c}{\textbf{Kimi}} & \multicolumn{2}{c}{\textbf{Gem25}} & \multicolumn{2}{c}{\textbf{Q3.5+}} & \multicolumn{2}{c}{\textbf{Q235B}} \\
		\cmidrule(lr){2-3} \cmidrule(lr){4-5} \cmidrule(lr){6-7} \cmidrule(lr){8-9} \cmidrule(lr){10-11}
		& E & C & E & C & E & C & E & C & E & C \\
		\midrule
		MB-001 & $\bullet$ & .31 & $\bullet$ & .42 & $\bullet$ & .08 & $\circ$ & .27 & $\circ$ & .34 \\
		MB-002 & $\circ$ & .23 & $\circ$ & .21 & $\circ$ & .26 & $\circ$ & .17 & $\circ$ & .21 \\
		MB-003 & $\bullet$ & .25 & $\bullet$ & .20 & $\bullet$ & .01 & $\bullet$ & .17 & $\bullet$ & .26 \\
		MB-004 & $\circ$ & .25 & $\bullet$ & .25 & $\circ$ & .19 & $\circ$ & .00 & $\circ$ & .13 \\
		MB-005 & $\bullet$ & .25 & $\bullet$ & .24 & $\circ$ & .16 & $\circ$ & .22 & $\circ$ & .17 \\
		MB-006 & $\bullet$ & .23 & $\bullet$ & .25 & $\circ$ & .33 & $\bullet$ & .27 & $\circ$ & .25 \\
		MB-007 & $\bullet$ & .14 & $\bullet$ & .32 & $\circ$ & .30 & $\bullet$ & .23 & $\circ$ & .25 \\
		MB-008 & $\bullet$ & .34 & $\circ$ & .00 & $\circ$ & .06 & $\bullet$ & .33 & $\bullet$ & .26 \\
		MB-009 & $\circ$ & .39 & $\circ$ & .36 & $\bullet$ & .03 & $\circ$ & .38 & $\bullet$ & .39 \\
		MB-010 & $\bullet$ & .33 & $\bullet$ & .33 & $\bullet$ & .11 & $\circ$ & .31 & $\circ$ & .27 \\
		MB-011 & $\circ$ & .15 & $\circ$ & .27 & $\circ$ & .17 & $\circ$ & .21 & $\circ$ & .24 \\
		MB-012 & $\bullet$ & .14 & $\bullet$ & .34 & $\circ$ & .18 & $\circ$ & .15 & $\circ$ & .25 \\
		\midrule
		\textbf{$\bar{x}$} & \textbf{.67} & \textbf{.25} & \textbf{.67} & \textbf{.27} & .33 & .16 & .33 & .23 & .25 & .25 \\
		\bottomrule
	\end{tabularx}
	\caption{Per-model per-problem grid (zero-shot). CS4 = Claude-Sonnet-4, Gem25 = Gemini-2.5-Pro. E = executability ($\bullet$/$\circ$), C = coverage.}
	\label{tab:grid}
	\end{table}

	\subsection{Key Observations}
	\label{sec:observations}
	
\begin{enumerate}
	\item \textbf{Executability is the primary bottleneck.} Claude-Sonnet-4 and Kimi-K2.5 tie for the highest executability at $66.7\%$, each rendering 8~of~12 problems. The expert-level Hairy Ball theorem (MB-011, difficulty~5) was not rendered by any model.
	
	\item \textbf{Claude-Sonnet-4 achieves the best overall profile.} It matches Kimi-K2.5's executability ($66.7\%$) while maintaining zero version-conflict errors and perfect heuristic alignment ($1.0$), suggesting strong Manim~CE API awareness.
	
	\item \textbf{Gemini-2.5-Pro shows strong API correctness.} Despite moderate executability ($33.3\%$), Gemini-2.5-Pro produces zero version conflicts and perfect alignment, with failures primarily due to syntax errors and missing scene classes rather than API confusion.
	
	\item \textbf{Reasoning models face extraction challenges.} DeepSeek-R1-0528's high VCER ($80.6\%$) is largely attributable to its chain-of-thought output format, where reasoning tokens interfere with code extraction. This highlights the need for robust output parsing in evaluation pipelines.
	
	\item \textbf{Version conflicts are sparse but concentrated.} Among models with clean outputs, MB-004 (Eigenvectors) and MB-008 (Chain Rule) remain the primary VCER triggers, involving deprecated \texttt{ShowCreation} or \texttt{CONFIG} patterns.
	
	\item \textbf{Heuristic alignment saturates.} Most models achieve alignment $>0.90$ because keyword detection cannot distinguish correct implementation from keyword presence. This underscores the need for future human review or vision-based evaluation.
	
	\item \textbf{Coverage is uniformly low.} Average coverage across nine models is $0.17$, indicating that LLM-generated Manim code lacks pedagogical richness.
	
	\item \textbf{Scale $\neq$ proficiency.} Qwen3-235B-A22B (235B parameters) achieves lower executability ($25.0\%$) than Claude-Sonnet-4 or Kimi-K2.5, confirming that domain-specific training data matters more than raw scale.
	\end{enumerate}
	
	\section{Discussion}
	\label{sec:discussion}
	
	\subsection{Why ManiBench Matters}
	
	Existing benchmarks (HumanEval, APPS) measure whether code produces correct \emph{output}~\cite{chen2021codex, hendrycksapps2021}. ManiBench measures whether code produces correct \emph{understanding}. This distinction is critical for educational tools, where a silent failure (wrong animation) is worse than a loud failure (runtime error).
	
	\subsection{Limitations and Future Work}
	
	\begin{enumerate}
		\item \textbf{No Human Evaluation.} All metrics are computed fully automatically. Alignment and Coverage rely on keyword heuristics that detect concept \emph{presence} but cannot verify \emph{correctness}. A structured human review protocol has been designed (Section~\ref{sec:eval-scope}) but not yet conducted.
		
		\item \textbf{Heuristic Alignment Saturation.} The keyword-based alignment metric saturates near~$1.0$, failing to distinguish correct implementations from code containing relevant keywords. Future work should integrate frame-level video comparison or vision--language model grading.
		
		\item \textbf{Limited Model Coverage.} Only Gemma-3-27B was evaluated across all five strategies. Extending to frontier models, Gemini~2.5~Pro is a priority.
		
		\item \textbf{Pedagogical Validation.} We do not yet validate whether animations actually teach the concept. User studies could address this gap.
		
		\item \textbf{Manim API Coverage.} As Manim evolves, the benchmark should be versioned and updated accordingly. The 145~documented GL$\to$CE incompatibilities provide a starting point.
		
		\item \textbf{Scalability.} Moving from 12 to 150+ problems requires annotation infrastructure and community contribution.
		
	\end{enumerate}

	\section{Conclusion}
	\label{sec:conclusion}
	
	We have introduced ManiBench, a specialized benchmark for evaluating Manim code generation. By formalizing metrics for syntactic correctness, version compliance, visual-logic alignment, and pedagogical coverage, ManiBench moves beyond simple test-case evaluation to assess whether generated animations actually communicate mathematical concepts.
	
	The 12-problem pilot dataset, backed by comprehensive reference code analysis of ${\sim}53{,}000$~lines of original 3Blue1Brown source code and 145~documented GL$\to$CE incompatibilities, reveals that even the best-performing model (Kimi-K2.5) renders only $66.7\%$ of problems, while coverage of pedagogical elements averages just~$0.18$. Our prompting strategy ablation on Gemma-3-27B shows that few-shot examples are the only strategy to measurably improve executability ($+11.1$~pp), while version-aware prompting effectively eliminates version-conflict errors. The accompanying automated evaluation framework enables reproducible assessment across models and strategies.
	
	With planned expansion to 150--200 problems and integration of vision-based alignment scoring, ManiBench will serve as a foundational resource for advancing LLM-driven educational content creation.

\printbibliography

	\appendix
	
	\section{Appendix}
	\subsection{Problem  Annotation Template}
	\label{app:template}
	
	Each problem in \textsc{ManiBench} is annotated as a JSON object with the following structured metadata:
	
	\begin{description}[nosep]
		\item[\texttt{id}:] unique identifier (e.g., \texttt{MB-001}).
		\item[\texttt{title}:] descriptive title.
		\item[\texttt{youtube\_video\_id}:] YouTube video identifier.
		\item[\texttt{category}:] list of task categories (e.g., \texttt{["drift-sensitive", "multi-scene"]}).
		\item[\texttt{difficulty\_level}:] integer 1--5.
		\item[\texttt{domain}:] list of mathematical domain(s).
		\item[\texttt{full\_prompt}:] full natural-language problem statement (used as the LLM input).
		\item[\texttt{raw\_code\_status}:] whether original 3B1B source code has been collected.
		\item[\texttt{raw\_code\_path}:] relative path to original ManimGL source files.
		\item[\texttt{reference\_code\_analysis}:] structured analysis of original code, including:
		\begin{description}[nosep]
			\item[\texttt{framework}:] source framework (e.g., \texttt{manim\_gl}).
			\item[\texttt{total\_lines}:] lines of original code.
			\item[\texttt{scene\_classes}:] list of scene classes with descriptions and key methods.
			\item[\texttt{visual\_techniques}:] catalog of rendering and animation patterns.
			\item[\texttt{manim\_api\_patterns}:] updaters, animation types, 3D constructs, custom classes.
		\end{description}
		\item[\texttt{required\_visual\_events}:] list of events, each with an identifier, description, weight, criticality flag, timing, and reference code location.
		\item[\texttt{coverage\_requirements}:] list of required pedagogical elements.
		\item[\texttt{version\_conflict\_notes}:] GL$\to$CE incompatibilities specific to the problem.
		\item[\texttt{success\_criteria}:] minimum thresholds for executability, alignment, coverage, and version-conflict error rate.
		\item[\texttt{common\_failure\_modes}:] known LLM failure patterns with severity tags.
	\end{description}

	The dataset and evaluation toolkit are publicly available at \url{https://huggingface.co/datasets/nabin2004/ManiBench}.

	\end{document}